\documentclass[twoside]{article} \usepackage{aistats2017}

\usepackage[numbers, square]{natbib}
\usepackage[utf8]{inputenc} 
\usepackage[T1]{fontenc}    
\usepackage{lmodern}
\usepackage{hyperref}       
\usepackage{url}            
\usepackage{booktabs}       
\usepackage{amsfonts}       
\usepackage{nicefrac}       
\usepackage{microtype}      
\usepackage{amsmath}
\usepackage{amssymb}
\usepackage{bbm}
\usepackage{algorithm,algorithmicx}
\usepackage[noend]{algpseudocode}
\usepackage{tikz}
\usepackage{pgfplots}
\usepackage{wrapfig}
\usepackage{caption}
\usepackage{subcaption}
\usepackage{multirow}
\usepackage{enumitem}

\usetikzlibrary{automata,topaths}
\newcommand*\Let[2]{\State #1 $\gets$ #2}

\algrenewcommand\algorithmicrequire{\textbf{Precondition:}}
\algrenewcommand\algorithmicensure{\textbf{Postcondition:}}

{\bfseries}{\rmfamily}

\newcommand\proofsymbol{\frame{\rule[0pt]{0pt}{8pt}\rule[0pt]{8pt}{0pt}}}
   {\trivlist\item[\hskip\labelsep{\it #1.}]}
   {\hspace*{\fill}\proofsymbol\endtrivlist}

\usepackage{xcolor}
\colorlet{darkgreen}{black!40!green}

\usepackage{alberpaper}
\createcommentshortcuts{max}{Max}
\createcommentshortcuts{jul}{Julian}
\createcommentshortcuts{mar}{Marius}

\begin{document}

%

%


\twocolumn[
  
\aistatstitle{Distributed Optimization of Multi-Class SVMs}

\aistatsauthor{  Maximilian Alber \textsuperscript{\normalfont 1} $\dagger$\\
  \And
  Julian Zimmert \textsuperscript{\normalfont 3} $\dagger$\\
  \And
  Urun Dogan \textsuperscript{\normalfont 2}\\
  \And
  Marius Kloft \textsuperscript{\normalfont 3}\\
}

\vspace*{0.2cm}
\aistatsaddress{
  \textsuperscript{1} Berlin Big Data Center \\
  Berlin Institute of Technology \\
  10587 Berlin, Germany
  \And
  \textsuperscript{2} Microsoft Research \\
  Cambridge CB1 2FB\\ United Kingdom
  \And
  \textsuperscript{3} Department of Computer Science \\
  Humboldt University of Berlin \\
  10099 Berlin, Germany
}

]

\begin{abstract}
\vspace{-0.3cm}
Training of one-vs.-rest SVMs can be parallelized over the number of classes in a straight forward way.
Given enough computational resources, one-vs.-rest SVMs can thus be trained on data involving a large number of classes.
The same cannot be stated, however, for the so-called all-in-one SVMs, 
which require solving a quadratic program of size quadratically in the number of classes.
We develop distributed algorithms for two all-in-one SVM formulations (Lee et al. and Weston and Watkins)
that parallelize the computation evenly over the number of classes.
This allows us to compare these models to one-vs.-rest SVMs on unprecedented scale.
The results indicate superior accuracy on text classification data.
\vspace{-0.3cm}
\end{abstract}

\section{Introduction}
\vspace{-0.2cm}

Modern data analysis requires computation with a large number of classes.
As examples, consider the following.
(1) We are continuously monitoring the internet for new webpages, which we would like to categorize. 
(2) We are collecting data from an online feed of photographs that we would like to classify into image categories.
(3) We add new articles to an online encyclopedia and intend to predict the categories of the articles.

The problems above
involve a large number of classes, typically at least in the thousands.
This motivates research on scaling up multi-class classification methods.
In the present work, we address scaling up multi-class support vector machines (MC-SVMs) \citep{vapnik:slt}.
There are two major types of MC-SVMs:
\begin{itemize}[leftmargin=*,nolistsep]
\vspace{-0.1cm}
\item[1.] One-vs.-one (OVO) and one-vs.-rest (OVR) MC-SVMs decompose the problem into multiple binary subproblems that are subsequently aggregated \citep{vapnik:slt,rifkin2004dov}.
  Training can be parallelized in a straight forward way.
\item[2.] \emph{All-in-one} MC-SVMs extend the concept of the margin to multiple classes.
Because there is no unique extension of the margin concept, multiple all-in-one MC-SVMs have been proposed, 
including the ones by Crammer and Singer (CS) \citep{crammer2002aim}, Lee, Lin, and Wahba (LLW) \citep{lee2004multicategory}, and Weston and Watkins (WW) \citep{weston1999svm,vapnik:slt}.
See \cite{rifkin2004dov,allwein2001reducing,hsu2002cmm, hill2007framework,liu:2007,guermeur:2007,UrunJMLR16} for conceptual and empirical comparisons.
\end{itemize}

Recently, Dogan et al. \cite{UrunJMLR16} have compared the various all-in-one MC-SVM variants on rather moderately sized datasets
and showed advantages of all-in-one MC-SVMs over OVR MC-SVM,
but --- so far --- slow training time has prohibited comparisons on data involving a large number of classes.

The reason is that (linear) state of the art solvers require time complexity of $\mathcal{O}(\bar{d} \bar{n}\cdot \mathcal{C}^2)$ 
and memory complexity at least of $\mathcal{O}(\bar{n}\mathcal{C}^2)$,
where $d$ is the feature dimensionality, $\bar{d}$ the average number of non-zeros ($\bar{d}=d$ for dense data),
 and $\bar{n}$ the average number of samples per class.
This quadratic dependence on the number of classes $\mathcal C$ can be prohibitive for large $\mathcal C$,
often leaving OVO and OVR as the only MC-SVM options in the big data setting.

In this paper, we develop distributed algorithms where up to $\mathcal{O}(\mathcal{C})$ nodes solve WW and LLW in parallel, 
dividing model and computation evenly.


The algorithm proposed for WW draws inspiration from a major result in graph theory: the solution to the 1-factorization problem of a graph \citep{bondy1976graph}.
The idea is that the optimization of a single coordinate $\alpha_{i,c}$ of the dual objective involves only the two hypotheses $w_{y_i}$ and $w_c$. 
As in the 1-factorization problem, we can thus form pairs of classes where the corresponding blocks of coordinates can be optimized in parallel.

On the other hand, we parallelize LLW training by introducing an auxiliary variable $\bar{w}$ into 
the dual problem that decouples the objective into a sum over $\mathcal C$ many independent subproblems. 

We provide both multi-core and distributed implementations of the proposed algorithms. 
We report on empirical runtime comparisons of the proposed solvers with the one-vs.-rest implementation by LIBLINEAR \citep{fan:2008}  
on text classification data taken from the LSHTC corpus \citep{LSHTC}.

The main contributions of this paper are the following:
\begin{itemize}
  \item We propose the first distributed, exact solver for WW and LLW.
	\item We provide both multi-core and truly distributed implementations of the solver.
  \item We give the first comparison of WW, LLW, and OVR on the DMOZ data from the LSHTC '10--'12 corpora using the full feature resolution.
\end{itemize}

We expect that the present work starts a line of research on parallelization of exact training of various all-in-one MC-SVMs,
including Crammer and Singer, multi-class maximum margin regression \cite{szedmak2006lvl}, and 
the reinforced multicategory SVM\citep{liu2011reinforced}, enabling comparison of all these methods on large datasets.

The paper is structured as follows. In the next section we discuss the problem setting and preliminaries. 
In Section~\ref{sec:algos}, we present the proposed distributed algorithms for LLW and WW, respectively.
We analyze their convergence empirically in Section~\ref{sec:experim}.
Section~\ref{sec:related} discusses related work and Section~\ref{sec:concl} concludes.

\section{Preliminaries}\label{sec:prelim}

We consider the following problem. We are given data $(x_1,y_1), \ldots,(x_n,y_n)$ with $x_i\in\mathbb{R}^d$ and $y_i\in\{1,...,\mathcal{C}\}$. 
Each class has in average $\bar{n}$ samples. The largest number of samples for a single class is $n_{max}$. 
We are predicting using the model
\begin{align}\label{eq:pred}
\hat{y}(x) := \argmax_c w_c^Tx ,
\end{align}
where $W=(w_1,..,w_\mathcal{C})\in\mathbb{R}^{d\times\mathcal{C}}$ are unknown parameters. 
The aim is to efficiently find good parameters in order to predict well on new data using \eqref{eq:pred}.

To address this problem setting, a number of generalizations of the binary SVM \citep{cortes1995support} have been proposed. 
We are specifically studying the following two formulations, dropping the bias terms in both.
Throughout this paper, $l(x) = \max\{0,1-x\}$ will denote the hinge-loss.

\paragraph{Lee, Lin, and Wahba (LLW) \citep{lee2004multicategory}}
\begin{equation}
\begin{aligned}
\min_{W}&\sum_{c=1}^\mathcal{C}\left[\frac{1}{2}||w_c||^2 + C\sum_{i:y_i\neq c}l(-w_c^Tx_i)\right]\\ \,\mbox{s.t. } & \sum_cw_c = 0 
\end{aligned}
\end{equation}

\paragraph{Weston and Watkins (WW) \citep{weston1999svm}}
\begin{equation}
\begin{aligned}
\min_{W}\sum_{c=1}^\mathcal{C}\left[\frac{1}{2}||w_c||^2 + C\sum_{i:y_i\neq c}l(w_{y_i}^Tx_i-w_c^Tx_i)\right]
\end{aligned}
\end{equation}

Both formulations lead to very similar dual problems, as shown below.  
For the dualization of WW, we refer to \cite{Keerthi2008}. 
The LLW dual is given in Appendix~\ref{app:dualization},
where we introduce an auxiliary variable $\bar{w}$ that is exploited by our distributed algorithm, 
as explained in the next section.

\vspace{-0.5cm}
\begin{align}\label{dual:LLW}
\tag{LLW}
\begin{split}
  & \max_{\alpha\in\mathbb{R}^{n\times\mathcal{C}},\bar{w}\in\mathbb{R}^d} \sum_{c=1}^\mathcal{C}\left[-\frac{1}{2}||-X\alpha_c+\bar{w}||^2 + \sum_{i:y_i\neq c}\alpha_{i,c}\right]\\
  & \qquad~ \text{s.t.} \qquad \forall i:~\alpha_{i,y_i}= 0,  \forall c\neq y_i: 0\leq\alpha_{i,c}\leq C 
\end{split}
\end{align}

\vspace{-0.2cm}
\begin{align}\label{dual:WW}
\tag{WW}
\begin{split}
  & \max_{\alpha\in\mathbb{R}^{n\times\mathcal{C}}} \quad \sum_{c=1}^\mathcal{C}\left[-\frac{1}{2}||-X\alpha_c||^2 + \sum_{i:y_i\neq c}\alpha_{i,c}\right] \\
  & \quad \text{s.t.} \qquad~ \forall i:~ \alpha_{i,y_i}=-\sum_{c:c\neq y_i}\alpha_{i,c},\\ & \quad \qquad~ \quad~ \forall c\neq y_i: ~ 0\leq\alpha_{i,c}\leq C 
\end{split}
\end{align}

\vspace{-0.1cm}
\section{Distributed Algorithms}\label{sec:algos}

In this section, we derive algorithms that solve (LLW) and (WW) in a distributed manner. With start by addressing LLW.

\vspace{-0.1cm}
\subsection{Algorithm for Lee, Lin, and Wahba}

First note the following optimality condition in (LLW): 
\begin{equation}\label{eq:w_mean}
  \bar{w}=\frac{1}{\mathcal C}\sum_{c=1}^\mathcal{C}X\alpha_c.
\end{equation}
Which was exploited by prevalent solvers to remove the variable $\bar{w}$ from the optimization,
yielding the objective 
$\textrm{obj}(\alpha)=\sum_{c=1}^\mathcal{C}\left[-\frac{1}{2}||-X\alpha_c+\bar{w}||^2 + \sum_{i:y_i\neq c}\alpha_{i,c}\right].$
In contrast, the core idea of our LLW solver is to actually keep this auxiliary variable, as it decouples the objective function into the following sum:
\begin{equation}\label{eq:decomp}
  \textrm{obj}(\alpha) = \sum_{c=1}^\mathcal{C} \textrm{subobj}(\alpha_c,\bar{w}).
\end{equation}

Then we perform dual block coordinate ascent (DBCA) \cite[][Algorithm 3.1]{Keerthi2008} with a specially tailored block structure,
considering as blocks $\bar{w}$ as well as each single coordinate $\alpha_{i,c}$.
As we observe from \eqref{eq:decomp}, the optimization of the columns $\alpha_{:,c}$ is mutually independent of each other, given fixed $\bar{w}$.
Hence, it can be distributed evenly over $\mathcal C$ many nodes. 
On the $c$th node, we run coordinate ascend on the subobjective $\textrm{subobj}(\alpha_c,\bar{w})$ over $\alpha_{i,c}, i=1,\ldots,n$,
as described in the next paragraph.
After one epoch of $\alpha$ computation, the variable $\bar{w}$ is updated via \eqref{eq:w_mean}.
The final algorithm is shown in Algorithm~\ref{alg:LWS}.



\begin{algorithm}[t]
  \begin{algorithmic}[1]
	\caption{Lee, Lin, and Wahba\label{alg:LWS}}
	\Function{solve-LLW}{$C,X,Y$}
	\For{$c=1..\mathcal{C}$}{\, \textbf{in parallel}}	
		\Let{$w_c$}{$0$}
		\Let{$\alpha_c$}{$0$}
		\For{$i\in I$}{}
			\Let{$k_i$}{$x_i^Tx_i$}
		\EndFor
      \While{\textbf{not} $\operatorname{optimal}$}
	\Let{$\operatorname{optimal}$}{True}
	\State shuffleData()
    \For{$i \in I\setminus I_c$}{}
		\State{solve1DimLLW($i,c$)}
	\EndFor
    \Let{$\bar{w}$}{ Reduce($\sum_cw_c/\mathcal{C}$)}
    \Let{$w_c$}{$w_c-\bar{w}$}
	\EndWhile
	\EndFor
    \EndFunction
  \end{algorithmic}
\end{algorithm}


As necessary step within Algorithm~\ref{alg:LWS}, we need to update every single $\alpha_{i,c}$.
Optimizing $\alpha_{i,c}$ is solving the problem
\begin{align}\label{eq:llw1dsubprob}
\begin{split}
\argmax_{\delta}D(\alpha + \delta e_{i,c},\bar{w})\\
\mbox{s.t. }0\leq \alpha_{i,c}+\delta \leq C,
\end{split}
\end{align}
where $e_{i,c}\in\mathbb R^{n\times\mathcal C}$ is one at the $(i,c)$th coordinate and zero elsewise.
Set $w_c:=-X\alpha_c+\bar{w}$; then the gradient for $\delta$ is 
$ \frac{\partial}{\partial\delta}\left[D(\alpha + \delta e_{i,c})\right]=x_i^Tw_c -x_i^Tx_i\delta +1.$
Hence, the optimal solution of \eqref{eq:llw1dsubprob} is given by
$
\delta = \min\{C-\alpha_{i,c}, \max\{-\alpha_{i,c} , -\frac{x_i^Tw_c -1}{x_i^Tx_i}\}\}.
$
The corresponding pseudo-code can be found in Algorithm~\ref{alg:LLWsub}.

  \begin{algorithm}[t]
   \caption{Solving 1-dim sub-problem of LLW \label{alg:LLWsub}}
  \begin{algorithmic}[1]
	\Function{solve1DimLLW}{$i$,$c$}
	\State{\textbf{global} $C,X,k,\alpha_c,w_c,\operatorname{optimal}$}
	\Let{$g$}{$w_c^Tx_i-1$}        
	\If{$g<-\epsilon$ and $\alpha_{i,c}<C$}
          \Let{$\delta$}{$\min\{C-\alpha_{i,c},-g/k_i\}$}
	\Let{$\operatorname{optimal}$}{False}
	\EndIf
	\If{$g>\epsilon$ and $\alpha_{i,c}>0$}
          \Let{$\delta$}{$\max\{-\alpha_{i,c},-g/k_i\}$}
	\Let{$\operatorname{optimal}$}{False}
	\EndIf
          \Let{$w_c$}{$w_c+\delta x_i$}
	\Let{$\alpha_{i,c}$}{$\alpha_{i,c}+\delta$}
\EndFunction
\end{algorithmic}
\end{algorithm}

\subsubsection{Convergence}\label{sec:conv}
  
It is well known that the block coordinate ascent method converges under suitable regularity conditions \cite[e.g.,][]{tseng2001convergence,bertsekas1995nonlinear}.
Our objective is continuously differentiable and strictly convex. 
The constraints are solely box constraints, hence the feasible set decomposes as a Cartesian product over the blocks.
Algorithm~\ref{alg:LWS} traverses the two blocks in cyclic order. 
Under these conditions, the DBCA method provably converges \cite[e.g,][Prop. 2.7.1]{bertsekas1995nonlinear}.

Note that in practice, we observed speedups by updating $\bar{w}$ in Algorithm~\ref{alg:LWS} after each tenth of an epoch, breaking the cyclic order. 
The blocks of coordinates are then traversed in so-called \emph{essentially cyclic order} \citep[e.g.,][Section 2]{tseng2001convergence},
meaning that there exists $T\in\mathbb N$ such that each block is traversed at least once after $T$ iterations.
Closer inspection of the proof in \cite[e.g,][Prop. 2.7.1]{bertsekas1995nonlinear} reveals that the result holds also under
this slightly more general assumption.

Further, we drop variables $\alpha_{i,c}$ in the optimization (shrinking) if they are not updated in three subsequent epochs.
Once the stopping condition holds, we run another epoch of optimization over all variables (including the ones that were dropped).
If the stopping criterion is then met, we terminate the algorithm and continue optimization elsewise.

\subsubsection{Implementation details}\label{sec:conv}

Our implementation uses OpenMPI for inter-machine \citep[MPI]{gropp1996high} and OpenMP \cite[MC]{dagum1998openmp} for intra-machine communication. 
Note that Algorithm~\ref{alg:LWS} has very mild communication requirements: the only communication needed is the sum of all weight vectors $\bar{w}=\sum_{c}w_c$.
Hence, MPI suffers very little from communication overhead between the various machines.
In practice, we may not be able to fully parallelize to the maximum of $\mathcal{C}$ cores;
therefore our algorithm will divide the set of classes into number-of-cores many chunks and optimize each class sequentially.

\subsection{Algorithm for Weston and Watkins}

In this section, we propose a distributed algorithm for WW,
which draws inspiration from the 1-factorization problem of a graph.
The approach is presented below.

\subsubsection{Preliminaries}\label{sec:wwdca}
\begin{algorithm}[t]
  \begin{algorithmic}[1]
	\Function{solve1DimWW}{$i$,$c$}
	\State{\textbf{global} $C,X,w_{y_i},w_c,\alpha_c,\operatorname{optimal}$}
	\Let{$g$}{$(w_{y_i}^T-w_c^T)x_i-1$}        
	\If{$g<-\epsilon$ and $\alpha_{i,c}<C$}
          \Let{$\delta$}{$\min\{C-\alpha_{i,c},-g/2k_i\}$}
	\Let{optimal}{False}
	\EndIf
	\If{$g>\epsilon$ and $\alpha_{i,c}>0$}
          \Let{$\delta$}{$\max\{-\alpha_{i,c},-g/2k_i\}$}
	\Let{optimal}{False}
	\EndIf
          \Let{$w_{y_i}$}{$w_{y_i}+\delta x_i$}
          \Let{$w_c$}{$w_c-\delta x_i$}
	\Let{$\alpha_{i,c}$}{$\alpha_{i,c}+\delta$} 
\EndFunction
\end{algorithmic}
\caption{ \label{alg:onedim} Solving 1-dim sub-problem of WW \label{alg:WWsub}}
\end{algorithm}

Our approach is based on running dual coordinate ascend \cite[][Algorithm 3.1]{Keerthi2008} over $\alpha_{i,c}$ on the (WW) objective function as follows.
Denote the objective in \eqref{dual:WW} by $D(\alpha)$ and recall from \cite{Keerthi2008} that optimizing $\alpha_{i,c}$ is solving the problem
\begin{align*}
\argmax_{\delta}D(\alpha + \delta \mathbf{e}_{i,c})\\
\mbox{s.t. }0\leq \alpha_{i,c}+\delta \leq C.
\end{align*}
Setting $w_c = \sum_{i:y_i\neq c}( -x_i\alpha_{i,c}+\sum_{c:c\neq y_i} x_i\alpha_{i,c}),$
the gradient for $\delta$ is given by $ \frac{\partial}{\partial \delta}\left[D(\alpha + \delta \mathbf{e}_{i,c})\right] = -x_i^T(w_{y_i}-w_c)-x_i^Tx_i\delta+1.$
Which is optimal at
\begin{align}\label{eq:opt_deltaWW}
\delta = \min\left(C-\alpha_{i,c}, \max\left(-\alpha_{i,c} ,\frac{x_i^T(w_{y_i}-w_c) -1}{2x_i^Tx_i}\right)\right).
\end{align}
This computation is summarized in Algorithm~\ref{alg:onedim}.

\subsubsection{Core Observation}
We observe from above that optimizing with regard to $\alpha_{i,c}$ will require only the weight vectors $w_{y_i}$ and $w_c$.
In other words, given four different classes $c_1,c_2,c_3,c_4$, the optimization of the block of variables 
$(\alpha_i,c_1)_{i:y_i=c_2}$---according to \eqref{eq:opt_deltaWW}---is independent of the optimization of the block $(\alpha_i,c_3)_{i:y_i=c_4}$. 
Hence it can be parallelized. 
In the next section we describe how we exploit this structure to derive a distributed optimization algorithm.

\subsubsection{Excursus: 1-Factorization of a Graph}\label{sec:1fac}

Assume that $\mathcal C$ is even. 
The core idea now is to form $\frac{\mathcal C}{2}$ many disjoint blocks $(\alpha_i,c_1)_{i:y_i=c_2},\ldots,(\alpha_i,c_{\mathcal C-1})_{i:y_i=\mathcal C}$ of variables.
Each of these blocks can be optimized in parallel.
The challenge is to derive a maximally distributed optimization schedule where  
each block $(\alpha_i,c_j)_{i:y_i=c_k}$ for any $j\neq k$ is optimized.

\begin{figure}[b]
\centering
\hspace{-0.0cm}
\scalebox{0.6}{
\begin{tikzpicture}[x=1cm,y=1cm]
\newcommand{\vertA}{(1.00000,0.00000),(0.62349,0.78183),(-0.22252,0.97493),(-0.90097,0.43388),(-0.90097,-0.43388),(-0.22252,-0.97493),(0.62349,-0.78183),(0,0)}
    \foreach \coord [count=\i] in \vertA {
      \coordinate [at=\coord, name=A\i];
      \ifnum\i<5
      \node [label={ [above] {360/7 * (\i -1)} : \i }] at \coord {\textbullet};
      \else
      \node [label={ [below] {360/7 * (\i -1)} : \i }] at \coord {\textbullet};
      \fi
    }
	\draw (A1) -- (A8);
	\draw (A2) -- (A7);
	\draw (A3) -- (A6);
	\draw (A4) -- (A5);
	\node at (0,-1.9) {$r=1$};
 \end{tikzpicture}
\begin{tikzpicture}[x=1cm,y=1cm]
\newcommand{\vertA}{(1.00000,0.00000),(0.62349,0.78183),(-0.22252,0.97493),(-0.90097,0.43388),(-0.90097,-0.43388),(-0.22252,-0.97493),(0.62349,-0.78183),(0,0)}
    \foreach \coord [count=\i] in \vertA {
      \coordinate [at=\coord, name=A\i];
      \ifnum\i<5
      \node [label={ [above] {360/7 * (\i -1)} : \i }] at \coord {\textbullet};
      \else
      \node [label={ [below] {360/7 * (\i -1)} : \i }] at \coord {\textbullet};
      \fi
    }
	\draw (A2) -- (A8);
	\draw (A1) -- (A3);
	\draw (A7) -- (A4);
	\draw (A6) -- (A5);
	\node at (0,-1.9) {$r=2$};
 \end{tikzpicture}
\begin{tikzpicture}[x=1cm,y=1cm]
\newcommand{\vertA}{(1.00000,0.00000),(0.62349,0.78183),(-0.22252,0.97493),(-0.90097,0.43388),(-0.90097,-0.43388),(-0.22252,-0.97493),(0.62349,-0.78183),(0,0)}
    \foreach \coord [count=\i] in \vertA {
      \coordinate [at=\coord, name=A\i];
      \ifnum\i<5
      \node [label={ [above] {360/7 * (\i -1)} : \i }] at \coord {\textbullet};
      \else
      \node [label={ [below] {360/7 * (\i -1)} : \i }] at \coord {\textbullet};
      \fi
    }
	\draw (A3) -- (A8);
	\draw (A2) -- (A4);
	\draw (A1) -- (A5);
	\draw (A6) -- (A7);
	\node at (0,-1.9) {$r=3$};
 \end{tikzpicture}
\hspace{0.6cm}
\begin{tikzpicture}[x=0.2cm,y=1.5cm]
	\node at (-1,0) {.};
	\node at (0,0) {.};
	\node at (1,0) {.};
 \end{tikzpicture}
\hspace{0.6cm}
\begin{tikzpicture}[x=1cm,y=1cm]
\newcommand{\vertA}{(1.00000,0.00000),(0.62349,0.78183),(-0.22252,0.97493),(-0.90097,0.43388),(-0.90097,-0.43388),(-0.22252,-0.97493),(0.62349,-0.78183),(0,0)}
    \foreach \coord [count=\i] in \vertA {
      \coordinate [at=\coord, name=A\i];
      \ifnum\i<5
      \node [label={ [above] {360/7 * (\i -1)} : \i }] at \coord {\textbullet};
      \else
      \node [label={ [below] {360/7 * (\i -1)} : \i }] at \coord {\textbullet};
      \fi
    }
	\draw (A7) -- (A8);
	\draw (A1) -- (A6);
	\draw (A2) -- (A5);
	\draw (A3) -- (A4);
	\node at (0,-1.9) {$r=7$};
\end{tikzpicture}
}
\caption{Illustration of the solution of the 1-factorization problem of a graph with $\mathcal C=8$ many nodes.\label{fig:1graph}
}
\end{figure}
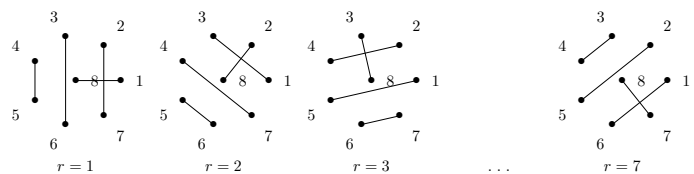

To better understand the problem, we consider the following analogy.
We are given a football league with $\mathcal C$ teams. 
Before the season, we have to decide on a schedule such that each team plays any other team exactly once.
Furthermore, all teams shall play on every matchday so that in total we need only $\mathcal C-1$ matchdays.
This problem is the \emph{1-factorization problem in graph theory} \citep[e.g.,][]{bondy1976graph}.  
The solution to this problem, illustrated in Figure~\ref{fig:1graph}, is as follows.

We arrange one node centrally and all other nodes in a regular polygon around the center node. 
At round $r$, we connect the centered node with node $r$ and connect all other nodes orthogonal to this line.
The pseudocode to compute the partner of a given node $c$ at a certain round $r$ is given in Algorithm~\ref{alg:matching} in the appendix.
Note that in case of an uneven number of classes, we introduce a dummy class $\mathcal C+1$, making the number of classes even.
We run the algorithm, but skip all computations involving the dummy class.

\subsubsection{Algorithm}

The algorithm, shown in Algorithm~\ref{alg:WWS}, 
performs DBCA over the variables $\alpha_{i,c}$ using the schedule derived in Section~\ref{sec:1fac}
and the coordinate updates derived in Section~\ref{sec:wwdca}.

    \begin{algorithm}[t]
      \caption{Watkins-Weston \label{alg:WWS}}
      \begin{algorithmic}[1]
   \Function{Solve-WW}{c,X,Y}
	
	\For{$c=1..\mathcal{C}$}{\, \textbf{in parallel}}	
	\Let{$w_c$}{$0$}
	\Let{$\alpha_c$}{$0$}
	\EndFor
	\For{$i\in I$}{}
		\Let{$k_i$}{$x_i^Tx_i$}
	\EndFor
	\While{\textbf{not} $\operatorname{optimal}$}
    \Let{$\operatorname{optimal}$}{True}
		\State shuffleData()
		\For{$r=1..\mathcal{C}-1$}
			\For{$c=1..\mathcal{C}$}{\, \textbf{in parallel}}
				\Let{$\tilde{c}$}{matchClass($\mathcal{C},c,r$)}
				\If{$\tilde{c}> c$}
					\For{$i\in I_c$}{}
						\State{solve1DimWW($i,\tilde{c}$)}
					\EndFor	
					\For{$i \in I_{\tilde{c}}$}{}
						\State{solve1DimWW($i,c$)}
					\EndFor	
				\EndIf
		  	\EndFor
		\EndFor
	\EndWhile
    \EndFunction
  \end{algorithmic}
    \end{algorithm}

\subsubsection{Convergence and Implementation details}

Furthermore, note that our algorithm performs the same coordinate updates as Algorithm 3.1 in \cite{Keerthi2008}.
Hence, they share the same favorable convergence behavior.
Formally, convergence is guaranteed for exactly the same reasons discussed in Section~\ref{sec:conv}.
We also employ the same speedup tricks, i.e. shrinking and updating every tenth of an epoch.

In practice, because of limitations of computational resources, we might not be able to fully parallelize to the maximum of $\mathcal{C}/2$ cores.
In that case, our algorithm divides the set of classes into number-of-cores many chunks and solves each bundle sequentially.
For optimal speedup, it is advisable to arrange the classes into chunks of equal number of classes and data points.

As with LLW, we implemented a mixed MPI-OpenMP solver for WW. 
However, note that, while LLW has mild communication needs, 
WW needs to pair the weight vectors of the matched classes $c$ and $\tilde{c}$ in each epoch,
for which $\mathcal{C}/2$ weight vectors needs to communicated among computers.
Therefore it is crucial to communicate efficiently.

We tackled the problem as follows. 
First of all, we use OpenMP for computations on a single machine (efficiently parallelizing among cores).
Here, due to the shared memory, no weight vectors need to be moved.
The more challenging task is to handle inter-machine communication efficiently.
Our approach is based on two key observations.

If the data is high-dimensional data, yet sparse, we keep the full weight matrix in memory for fast access, 
yet communicating only the non-zero entries between computers. 
Regardless of the increased computational effort, this takes only a fraction of time compared to sending the dense data.

Furthermore, we relax the WW matching scheme. Coming back to a football, consider each country hosts a league, and inside the league,
we match the teams as known. Now we would like to match teams across leagues. In order to do so, we first match the countries
with the scheme from Section~\ref{sec:1fac}. For each pair of countries, call them A and B,
every team from country A plays any other team from country B.
Coming back to classes and machines, this means we transfer bundles of classes (countries) between computers.
This drastically reduces the network communication.

\section{Experiments}\label{sec:experim}

This section is structured as follows. First we empirically verify the soundness of the proposed algorithms.
Then we introduce the employed datasets, on which we investigate the convergence and runtime behavior of the proposed algorithms
as well as the induced classification performance.

For the experimental setup we refer to \ref{sec:exprsetup} in the supplement. 
Each training algorithm was run three times, using randomly shuffled data, and the results were averaged.
Note that the training set is the same in each run,
but the different order of data points can impact the runtime of the algorithms.

\subsection{Validation of solver}\label{sec:validationexpr}

  \begin{table*}[t]
    \footnotesize
    \centering
    \caption{ {\normalsize Error on the test set and density in \% of the Shark solver (denoted S) and the proposed solver (denoted D). The complete table can be found in the supplement.}\label{tab:smallDataResults} }
    \begin{tabular}{ r | r r r r | r r r r}
    & \multicolumn{4}{c|}{\textbf{\textit{Error}}} & \multicolumn{4}{c}{\textbf{\textit{Model-Density}}} \\
    \textbf{Dataset:}  & \textbf{D-LLW} & \textbf{S-LLW} & \textbf{D-WW} & \textbf{S-WW} & \textbf{D-LLW} & \textbf{S-LLW} & \textbf{D-WW} & \textbf{S-WW} \\ 
    \hline

    \textbf{news20} & & & & & & & & \\
    \textit{$\log(C)$: -1} & 29.23 & 29.23 & 15.32 & 15.30   & 97.24 & 97.24 & 51.16 & 49.72 \\
    \textit{    0}         & 22.97 & 22.97 & 14.80 & 14.80   & 97.24 & 97.24 & 44.74 & 42.70 \\
    \textit{    1}         & 16.15 & 16.15 & 15.98 & 15.98   & 97.17 & 97.04 & 45.97 & 43.47 \\
    
    \textbf{rcv1} & & & & & & & & \\
    \textit{$\log(C)$: -1} & 47.96 & 47.96 & 11.31 & 11.31   & 78.00 & 78.00 & 26.42 & 23.45 \\
    \textit{    0}         & 33.27 & 33.41 & 11.52 & 11.52   & 78.00 & 77.98 & 22.93 & 20.12 \\
    \textit{    1}         & 12.03 & 12.03 & 12.03 & 12.03   & 78.00 & 77.98 & 23.05 & 20.06 \\

  \end{tabular}
  \end{table*}

In our first experiment, we validate the correctness of the proposed solvers. 
We downloaded data from the \textit{LIBLINEAR} \cite{fan:2008}\footnote{\url{https://www.csie.ntu.edu.tw/~cjlin/liblinear/}} and UCI \cite{asuncion2007uci}\footnote{\url{https://archive.ics.uci.edu/ml/datasets.html}} dataset repositories.
Where training and test splits are unavailable, we split the data once into $90\%$ train and $10\%$ test sets.
For each dataset, the optimal feature scaling was selected, in order to maximize the average accuracy on the test sets.
Datapoints in \emph{iris} and \emph{news} were thus normalized to unit norm, \emph{letter} and \emph{satimage} were normalized to unit variance. 
All other data was considered unnormalized. 

Then we compare our LLW and WW solvers with the state-of-the-art implementation contained in the ML library shark \citep{igel2008s}.
We implemented the same stopping criteria as \cite{igel2008s}.
The full results (averaged over 10 runs) are shown in Table~\ref{tab:smallDataResultsAll} in the supplement. A subset is given in Table~\ref{tab:smallDataResults}.
We observe good accordance of the results and model sparsity of the proposed solvers and the reference implementation from the Shark toolbox,
thus confirming that our respective solvers are indeed exact solvers of LLW and WW.

At random we tested whether the duality-gap closes or not. We did this for both solvers with different $C$ values and datasets.
In any case the duality gap closed, i.e. decreased by an order of two magnitudes. 
Based on this we chose our stopping criteria $\epsilon$ equal to 0.1 for the LSHTC datasets.

\subsection{Datasets}

\begin{table}[h]
\vspace{-0.15cm}
\footnotesize
\centering
\setlength{\tabcolsep}{3pt}
\caption{The used datasets and their properties.}
\begin{tabular}{l|r r r r}
  \textbf{Dataset} &  \textbf{$\mathbf{n}$ train} &  \textbf{$\mathbf{n}$ test} & $\mathbf{\mathcal{C}}$ &  $\mathbf{d}$\\
  \hline
  \textbf{LSHTC-small} &  4,463 & 1,858  & 1,139 & 51,033\\
  \textbf{LSHTC-large} &  128,710 & 34,880  & 12,294 & 381,581\\
  \textbf{LSHTC-2012} &  383,408 & 103,435  & 11,947 & 575,555\\
  \textbf{LSHTC-2011} &  \;\;394,754 & \;\;104,263  & \;\;27,875 & \;\;594,158\\
\end{tabular}
\label{tbl:datasets}
\end{table}

We experiment on large classification datasets, where the number of classes ranges between 451 and 27,875. 
The relevant statistics of the datasets are shown in \ref{tbl:datasets}. The LSHTC-* datasets are high-dimensional text datasets taken from the well-known LSHTC corpus \citep{LSHTC}.
The datasets belong to the released competition rounds 1 to 3, i.e. '10-'12. LSHTC-2011 and LSHTC-2012 originate from the DMOZ corpus.
The features were extracted using TF/IDF representation and we use the full feature resolution for training.

\subsection{Speedup}

\begin{figure}
\pgfplotstableread{
x  LLW1R    LLW10R   WW1R    WW10R
1  1.00     1.00     nan     1.0000
2  1.71     1.43     nan     1.04
4  3.19     2.69     nan     1.73
8  5.90     4.78     nan     2.92
16  9.23     8.77     nan     4.81
32  nan      nan      nan     nan
}\datatableDSLa
\pgfplotstableread{
x  LLW1R    LLW10R   WW1R    WW10R
1  1.00     nan     nan     nan
2  1.97     1.94     nan     1.83
4  nan      2.85     nan     1.91
8  nan      5.21     nan     3.09
16  nan      8.99     nan     4.80
32  nan      16.45    nan     7.01
}\datatableDSLampia
\pgfplotstableread{
x  LLW1R    LLW10R   WW1R    WW10R
1  1.00     nan      nan     nan
2  1.97     nan      nan     nan
4  nan      3.67     nan     3.25
8  nan      5.42     nan     3.46
16  nan      10.21    nan     5.20
32  nan      18.10    nan     7.49
}\datatableDSLampib
\pgfplotstableread{
x  LLW1R    LLW10R   WW1R    WW10R
1  1.0000   1.0000   1.0000  1.0000
2  1.54     1.43     1.40    1.45
4  3.10     2.71     2.14    2.35
8  6.19     4.85     4.27    3.22
16 11.99    8.95     6.66    3.88
32 nan      nan      nan  nan
}\datatableDSLb
\pgfplotstableread{
x  LLW1R    LLW10R   WW1R    WW10R
1  1.0000   1.0000   1.0000  1.0000
2  nan      1.96     nan     1.53
4  nan      2.81     nan     2.08
8  nan      5.23     nan     3.37
16 nan      9.22     nan     5.13
32 nan      16.82    nan     6.7
}\datatableDSLbmpi

\centering
\hspace{-1cm}

\begin{tikzpicture}[scale=0.8]
  \begin{axis}[ 
      legend style={at={(0.05,1.0)},anchor=north west},
      title={\textbf{LSHTC-large}},
      xlabel={Number of Nodes},
      xmode=log,
      log basis x=2,
      xmin=0.9,
      xmax=35,
      xtick={1, 2, 4, 8, 16, 32},
      ylabel={Speedup},
      ymode=log,
      log basis y=10,
      ymin=0.9,
    ]
    \addplot [color=red, solid, mark=*, mark options={solid}] table [y=LLW10R] {\datatableDSLa};
    \addplot [color=red, dashed, mark=square*, mark options={solid}] table [y=LLW10R] {\datatableDSLampia};
    \addplot [color=red, dotted, mark=triangle*, mark options={solid}] table [y=LLW10R] {\datatableDSLampib};
    \addplot [color=blue, solid, mark=*, mark options={solid}] table [y=WW10R] {\datatableDSLa};
    \addplot [color=blue, dashed, mark=square*, mark options={solid}] table [y=WW10R] {\datatableDSLampia};
    \addplot [color=blue, dotted, mark=triangle*, mark options={solid}] table [y=WW10R] {\datatableDSLampib};
    \legend{LLW-MC, LLW-MPI-2, LLW-MPI-4, WW-MC, WW-MPI-2, WW-MPI-4};
  \end{axis}
\end{tikzpicture}


\caption{Speed-up averaged over 10 repetitions respectively in the number of cores.\label{fig:speedup}}
\end{figure}
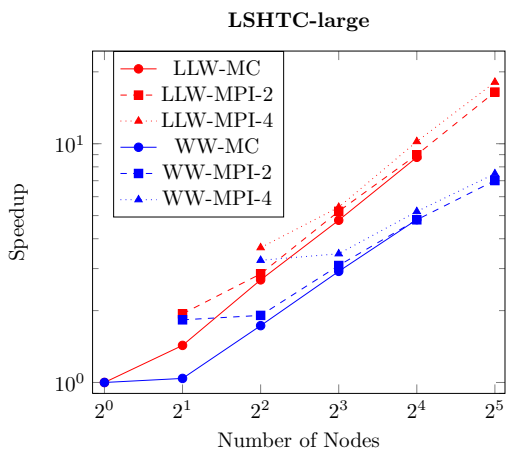

In order to measure the speedup provided by increasing the number of machines/cores, we run a fix amount of iterations over the whole LSHTC-large dataset.
We use 10 runs
over 10 iterations with a fixed parameter C equal $1$ without shrinking. While the MC execution works on one machine, the MPI executes on
two or four machines, i.e. spreading the used cores evenly on each node.

The results are shown in Figure \ref{fig:speedup}. LLW exhibits linear speedup regardless if distributed or not, due to the small communication cost.

For less than 4 cores WW has a similar, linear speedup. Then we assume that the cache bound
is reached\footnote{The used machines have two cpu-socket, i.e. two major caches.}.
When distributing on two machines (MPI), i.e doubling the cache-throughput,
despite the communication cost a higher speedup is reached.
This again vanishes as more cpus per machine are used.
This confirms our assumption and
suggests that on a system with more cache capacity a better speedup can be reached.



\subsection{Timing and Classification Results}
\label{sec:lshtcresults}

  Now we evaluate and compare the proposed algorithms on the LSHTC datasets for a range of C values, 
	i.e. we perform no cross-validation. For comparison we use solvers from the well-known \textit{LIBLINEAR} package, 
	namely the multi-core implementation with L2L1-loss (OVR, \cite{chiangparallel}) and the Crammer-Singer implementation (CS, \cite{fan:2008}).
    For the multi-core solvers, i.e. OVR and WW-MC, we use 16 cores. MPI spreads over 2 or 4 machines 
    using 8 and 4 cores respectively at each node, \textit{thus trains the model distributed}. Table~\ref{tab:lshtcResults} shows the error and the model sparsity for the compared solutions.
	We further provide the Micro-F1 and Macro-F1 score in Table~\ref{tab:lshtcF1Results} in the supplement.

  \begin{table*}[t]
    \footnotesize
    \centering
    \caption{Test set error and model density (in \%) as achieved by the OVR, CS, WW, and LLW solvers on the LSHTC datasets. For each solver the result with the best error is in bold font. For LLW entries with a '*' did not converge within a day of runtime.}\label{tab:lshtcResults}
    \begin{tabular}{ r | r r r r || r r r r }
    & \multicolumn{4}{c||}{\textbf{\textit{Error}}} & \multicolumn{4}{c}{\textbf{\textit{Model-Density}}} \\
    \textbf{Dataset:} & \textbf{OVR} & \textbf{CS} & \textbf{WW} & \textbf{LLW} & \textbf{OVR} & \textbf{CS} & \textbf{WW} & \textbf{LLW}  \\ 
    \hline
    
    \textbf{LSHTC-small}  & & & & & & & & \\
    \textit{$\log(C)$: -3} & 93.00 & 59.74 & 72.82 & \textbf{93.00}   & 92.74 & 11.11 & 69.73 & \textbf{92.74} \\
    \textit{   -2}         & 85.36 & 59.74 & 65.34 & 93.00   & 81.54 & 11.13 & 16.44 & 92.74 \\
    \textit{   -1}         & 74.54 & 59.74 & 57.59 & 93.00   & 46.76 & 11.12 &  6.06 & 92.74 \\
    \textit{    0}         & 64.37 & 55.49 & 54.57 & 93.00   & 38.20 & 11.76 &  5.74 & 92.74 \\
    \textit{    1}         & \textbf{57.75} & \textbf{54.57} & \textbf{54.41} & 93.00   & \textbf{38.63} & \textbf{11.69} &  \textbf{5.73} & 92.74 \\
 
    \textbf{LSHTC-large}  & & & & & & & & \\
    \textit{$\log(C)$: -3} & 88.12 & 58.57 & 66.47 & \textbf{95.86}   & 75.26 &  2.53 & 18.50 & \textbf{100.0} \\
    \textit{   -2}         & 85.21 & 58.57 & 60.58 & 95.86   & 45.14 &  2.53 &  4.45 & 100.0 \\
    \textit{   -1}         & 77.96 & 57.82 & 55.28 & 95.86   & 25.28 &  2.55 &  1.71 & 100.0 \\
    \textit{    0}         & 63.11 & \textbf{53.61} & \textbf{53.98} & 95.86   & 18.33 & \textbf{ 2.69} & \textbf{ 1.61} & 100.0 \\
    \textit{    1}         & \textbf{57.18} & 54.18 & 54.41 &     *   & \textbf{18.55} &  2.67 &  1.66 &     * \\

    \textbf{LSHTC-2012}  & & & & & & & & \\
    \textit{$\log(C)$: -3} & 83.66 & 49.81 & 58.02 & \textbf{92.63}   & 72.60 &  1.73 & 16.97 & \textbf{99.52} \\
    \textit{   -2}         & 75.15 & 49.65 & 50.20 & 92.63   & 46.20 &  1.71 &  4.06 & 99.52 \\
    \textit{   -1}         & 60.38 & 46.14 & 44.94 & 92.63   & 25.87 &  1.76 &  1.52 & 99.52 \\
    \textit{    0}         & 47.33 & \textbf{42.67} & \textbf{44.01} &     *   & 18.20 &  \textbf{2.06} & \textbf{ 1.42} &     * \\
    \textit{    1}         & \textbf{46.83} & 45.60 & 46.15 &     *   & \textbf{18.46} &  2.09 &  1.47 &     * \\

    \textbf{LSHTC-2011}  & & & & & & & & \\
    \textit{$\log(C)$: -3} & 87.95 & 59.09 & 68.19 & \textbf{96.18}   & 72.38 &  1.57 & 13.49 & \textbf{100.0} \\
    \textit{   -2}         & 85.85 & 59.09 & 62.14 & 96.18   & 45.97 &  1.57 &  3.16 & 100.0 \\
    \textit{   -1}         & 76.78 & 58.18 & 57.31 & 96.18   & 25.97 &  1.55 &  1.19 & 100.0 \\
    \textit{    0}         & 63.11 & \textbf{55.58} & \textbf{56.94} &     *   & 18.24 & \textbf{ 1.69} & \textbf{ 1.11} &     * \\
    \textit{    1}         & \textbf{60.01} & 57.78 & 58.32 &     *   & \textbf{18.46} &  1.70 &  1.14 &     * \\
    
  \end{tabular}
  \end{table*}

  For all datasets the canonical multi-class formulations, i.e. CS and WW, perform significantly better than OVR. 
	On one hand the error is smaller and the F1-scores better. On the other hand the learned models are much sparser, 
	i.e. up to a magnitude. The results justify the increased solution complexity of canonical formulations.

  Comparing CS and WW, CS performs as well or slightly better at classifying. Though WW leads to a sparser model. 
	To the best of our knowledge this is the first comparison of these well-known multi-class SVMs on the studied LSHTC data.

\begin{figure*}
\pgfplotstableread{
x  LL CS WWMC WWMPIL WWMPID2 WWMPID4
-2 4072.47 1311.25 534.61 646.57  897.32  949.79
-1 3024.74 2267.25 801.65 937.73  1046.06 955.94
0 3097.85 3512.93 1078.32 1558.17 1623.74 1610.36
1 3099.62 4659.54 1379.97 2216.13 2681.47 3016.48
}\datatableDSLSHTCa
\pgfplotstableread{
x  LL CS WWMC WWMPI2 WWMPI4
-2 28825.69 11277.34 3485.23 5963.24  4699.54
-1 24771.76 17609.56 6508.54 6882.01  6240.57
0 23894.07 26341.08 11361.44 12516.90 12522.83
1 25600.86 32904.49 21257.55 26354.00 27585.64
}\datatableDSLSHTCb
\centering
\hspace{-1cm}
\begin{tikzpicture}[scale=0.8]
  \begin{axis}[ 
      legend style={at={(-0.7,0.7)},anchor=north west},
      title={\textbf{LSHTC-large}},
      xlabel={$log(C)$},
      ylabel={Time [s]},
      xtick=data
    ] 
    \addplot [color=red, dashed, mark=*] table [y=LL] {\datatableDSLSHTCa};
    \addplot [color=red, mark=square*, mark options={solid}] table [y=CS] {\datatableDSLSHTCa};
    \addplot [color=blue, mark=square*, mark options={solid}] table [y=WWMC] {\datatableDSLSHTCa};
    \addplot [color=green, dashed, mark=square*, mark options={solid}] table [y=WWMPID2] {\datatableDSLSHTCa};
    \addplot [color=darkgreen, dashed, mark=*, mark options={solid}] table [y=WWMPID4] {\datatableDSLSHTCa};
    \legend{OVR,CS,WW-MP,WW-MPI-2, WW-MPI-4, WW-MPIL};
  \end{axis}
\end{tikzpicture}
\begin{tikzpicture}[scale=0.8]
  \begin{axis}[ 
      title={\textbf{LSHTC-2011}},
      xlabel={$log(C)$},
      ylabel={Time [s]},
      xtick=data
    ] 
    \addplot [color=red, dashed, mark=*] table [y=LL] {\datatableDSLSHTCb};
    \addplot [color=red, mark=square*, mark options={solid}] table [y=CS] {\datatableDSLSHTCb};
    \addplot [color=blue, mark=square*, mark options={solid}] table [y=WWMC] {\datatableDSLSHTCb};
    \addplot [color=green, dashed, mark=square*, mark options={solid}] table [y=WWMPI2] {\datatableDSLSHTCb};
    \addplot [color=darkgreen, dashed, mark=*, mark options={solid}] table [y=WWMPI4] {\datatableDSLSHTCb};
  \end{axis}
\end{tikzpicture}
\caption{Training time averaged over 10 repetitions per C. \label{fig:timing}}
\end{figure*}
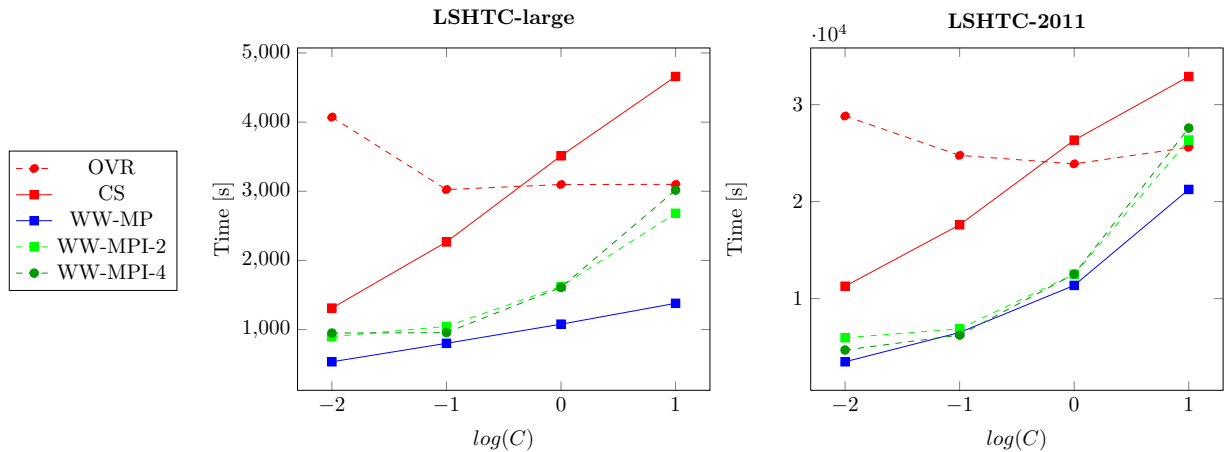

    From Figure~\ref{fig:timing},
    we observe that the runtime of our solver outperforms the one of OVR and CS by up to two orders of magnitude.
    Even when distributed our solver outperforms multi-core OVR in all except one case.

    All WW experiments use the same amount of cores, but with a varying degree of distribution.
    We observe that the communication imposes a modest overhead.

    Figure~\ref{fig:timing} confirms our assumption from the previous section.
    First, please note that shrinking reduces the active training set,
    i.e. reduces the computational effort and puts stress on the communication overhead.
    Therefore WW-MPI is regardless the higher speedup (see Figure~\ref{fig:speedup})
    slower than the multi-core version.
    Yet on a computationally intensive dataset as LSHTC-2011 the higher speedup
    cancels the overhead due to the communication for C equal to $0.1$ and $1$.

\subsubsection{Lin, Lee, \& Wahba}

  \begin{table}[t]
    \footnotesize
    \centering
    \caption{ {\normalsize Error, Micro-F1, and Macro-F1 on the test set and model density in \%  of the LLW solver on the LSHTC-small dataset.} \label{tab:LLWlargeC}}
    \begin{tabular}{ l | r r r}
    $\log(C)$: & \textbf{2} & \textbf{3} & \textbf{4} \\
    \hline
    \textbf{Error:} & \;\;87.73 & \;\;66.74 & \;\;59.31\\
    \textbf{Micro F1:} & 2.08 & 15.07 &  40.69\\
    \textbf{Macro F1:} & 12.27 & 33.26 &  24.58\\
    \hline
    \textbf{$W$-Density:} & 92.74 & 92.74 & 92.74\\
    \textbf{$\alpha$-Density:} & 99.88 & 99.87 & 99.90 \\
  \end{tabular}
  \end{table}

Knowing that LLW converges to the correct solution, as the duality-gap closes, the results indicate that the chosen C range is not suitable. 
For LSHTC-small we conducted experiments with much larger C values. And indeed, as shown in Table~\ref{tab:LLWlargeC}, LLW performs best in a 
nearly unconstrained setting. In any case, the model learned by LLW is never sparse. Not in the weight matrix $W$, nor in dual factors $\alpha$. 
Resource limitations and slow convergence properties hindered us to conduct experiments with even larger $C$ values. It is left to future work to explore this space or even develop a unconstrained version of LLW.

\section{Discussion of Related Work}\label{sec:related}


Most approaches to parallelization of MCSVM training are based on OVO or OVR \cite{babbar2016tersesvm},
including a number of approaches that attempt to learn a hierarchy of labels 
\cite{bengio2010label,deng2011fast,gao2011discriminative,NIPS2015_5937,xiao2011hierarchical,gopal2013recursive}
or train ensembles of SVMs on individual subsets of the data \cite{Govada2015,govada2015novel,Lodi10,jenssen2012scatter}.

There is a line of research on parallelizing stochastic gradient (SGD) based training of MC-SVMs over multiple computers \cite{gupta2014training,do2014parallel}.
SGD builds on iteratively approximating the loss term by one that is based on a subset of the data (mini-batch).
In contrast, batch solvers (such as the ones proposed in the present paper) are based on the full sample.
In this sense, our approach is completely different to SGD.
While there is a long ongoing discussion whether the batch or the SGD approach is superior,
the common opinion is that SGD has its advantages in the early phase of the optimization, 
while classical batch solvers shine in the later phase.
In this sense, the two approaches are complementary and could also be combined.

The related work that is the most closest to the present work is by \cite{han2012dcmsvm}.
They build on the alternating direction method of multipliers (ADMM) \cite{boyd2011distributed}
to break the Crammer and Singer optimization problem into smaller parts,
which can be solved individually on different computers. 
In contrast to our approach, the optimization problem is parallelized over the samples, not the optimization variables.
In our problem setting, high-dimensional sparse data, the model size is vary large.
Because each node holds the whole model in memory,
this algorithm hardly scales with large label spaces.
E.g. consider Table~\ref{tbl:datasets}; the model for LSHTC-2011 contains $\approx 16*10^9$ parameters.
Note also that it is unclear at this point whether the approach of \cite{han2012dcmsvm} could be adapted to LLW and WW,
which are the object of study in the present paper.

Note that beyond SVMs there is a large body of work on distributed multi-class \cite[e.g.,][]{vw,gopal2013distributed} and multi-label learning algorithms \cite{prabhu2014fastxml},
which is outside of the scope of the present paper.
\section{Conclusion}\label{sec:concl}

We proposed distributed algorithms for solving the multi-class SVM formulations by
Lee et al. (LLW) and Weston and Watkins (WW).
The  algorithm addressing LLW takes advantage of an auxiliary variable,
while our approach to optimizing WW in parallel is based on the 1-factorization problem from graph theory.

The experiments confirmed the correctness of the solver (in the sense of an exact solver)
and show linear speedup when the number of cores is increased.
This speedup allows us to train LLW and WW on LSHTC datasets, for which results were lacking in the literature.

Our analysis contributed to comparing MC-SVM formulations on rather large data sets, where comparisons
were still lacking.
In comparison to OVR we showed that WW can achieve competitive classification results in less time,
while still leading to a much sparser model.
Unexpectedly, LLW shows clear disadvantages over the other MC-SVMs. 
Yet the favorable scaling properties make further research interesting, for instance regarding the development of an unconstrained algorithm.
We ease further research by publishing the source code under \url{https://github.com/albermax/xcsvm}.

Overcoming the limitations of a single machine, i.e. distribution, is a key problem and a key enabler in large scale learning.
To best of our knowledge, we are the \textit{first to train an exact, all-in-one MC-SVMs in a distributed manner}.
We hope this first step inspires further research in this context.

In the future, we would like to study extensions of the concepts presented in this paper to various more MC-SVMs, 
including the Crammer and Singer MC-SVM \citep{crammer2002learnability}, multi-class maximum margin regression \cite{szedmak2006lvl}, and 
the reinforced multicategory SVM\citep{liu2011reinforced}.

\subsubsection*{Acknowledgments}

We thank Rohit Babbar, Shinichi Nakajima, and Klaus-Robert Müller for helpful discussions.
We thank Giancarlo Kerg for pointing us to the graph 1-factorization problem.
We thank Ioannis Partalas for help regarding the LSHTC datasets.
MK acknowledges support by the German Research Foundation (DFG) under KL 2698/2-1 and by the Federal Ministry of Education and Research (BMBF) under 031L0023A and 031B0187B.
MA acknowledges support by the Federal Ministry of Education and Research (BMBF) under 01IS14013A.

\begin{small}
\bibliography{xmc}
\bibliographystyle{ieeetr}
\end{small}

\appendix

\clearpage
\part*{Supplementary Material}
\numberwithin{equation}{section}
\numberwithin{theorem}{section}
\numberwithin{figure}{section}
\numberwithin{table}{section}
\renewcommand{\thesection}{{\Alph{section}}}
\renewcommand{\thesubsection}{\Alph{section}.\arabic{subsection}}
\renewcommand{\thesubsubsection}{\Roman{section}.\arabic{subsection}.\arabic{subsubsection}}
\setcounter{secnumdepth}{-1}
\setcounter{secnumdepth}{3}

\section{Derivation of Lagrandgian Dual Problems}\label{app:dualization}

\subsection{Lin, Lee, and Wahba}
Using slack variables, the primal LLW problem reads
\begin{equation*}
\begin{aligned}
\min_{W}&\quad \sum_{c=1}^\mathcal{C}\left[\frac{1}{2}||w_c||^2 + C\sum_{i:y_i\neq c}\xi_{i,c}\right]\\
\mbox{s.t. }&\quad \sum_cw_c = 0 \\
   &\quad~~\forall i:~ \begin{matrix} \xi_{i,c} \geq 1+w_c^Tx_i \quad \\[3pt]
                \forall c\neq y_i:~~ \xi_{i,c}\geq 0 .
\end{matrix}
\end{aligned}
\end{equation*}
We introduce Lagrangian multipliers $\alpha\in\mathbb R^{n\times\mathcal C}$, $\beta\in\mathbb R^n$, and $\bar{w}\in\mathbb R^d$ with $\alpha_{i,c},\beta_i\geq0$.
The Lagrangian is given in Figure~\ref{fig:LLWLag}.
\begin{figure*}[t]
\begin{align*}
L(W,\xi,\alpha,\beta,\bar{w}) = \sum_{c=1}^\mathcal{C}&\left[\frac{1}{2}||w_c||^2 + \sum_{i:y_i\neq c}\left(C\xi_{i,c} + \alpha_{i,c}(1+w_c^Tx_i-\xi_{i,c})-\beta_{i,c}\xi_{i,c}\right)\right] -\bar{w}^T(\sum_cw_c).
\end{align*}
\caption{Lagrangian for LLW.\label{fig:LLWLag}}
\end{figure*}

Slater's condition holds; therefore, we have strong duality and can use the dual
\begin{align*}
\max_{\alpha,\beta,\bar{\alpha}}&\min_{W,\xi}~~L(W,\xi,\alpha,\beta,\bar{w})\\
\mbox{s.t. }&\quad \forall\,i ~ \forall\,c: ~ \alpha_{i,c},\beta_{i,c}\geq 0 .
\end{align*}
The partial derivatives are given by
\begin{align*}
&\frac{\partial}{\partial \xi_{i,c}}L(W,\xi,\alpha,\beta,\bar{w}) = C-\alpha_{i,c}-\beta_{i,c}\\
&\frac{\partial}{\partial w_c}L(W,\xi,\alpha,\beta,\bar{w}) = w_c +\sum_{i:y_i\neq c}\alpha_{i,c}x_i+\bar{w}.
\end{align*}
Setting those to zero leads to
\begin{align*}\label{eq:LLWweights}
\forall\,i\,\forall\,c:~~ & 0\leq \alpha_{i,c}\leq C \\[3pt]
&w_c = -\sum_{i:y_i\neq c}\alpha_{i,c}x_i+\bar{w}\\
& \quad = -X\alpha_c +\bar{w}.
\end{align*}
And plugging in into the lagrangian, finally gives the dual
\begin{align*}
\max_{\alpha\in\mathbb{R}^{n\times\mathcal{C}},\bar{w}\in\mathbb{R}^d} \quad \sum_{c=1}^\mathcal{C}\left[-\frac{1}{2}||-X\alpha_c+\bar{w}||^2 + \sum_{i:y_i\neq c}\alpha_{i,c}\right]\\
\end{align*}\vspace{-20pt}
\begin{equation}
\forall i: \quad 
\begin{aligned}
&\alpha_{i,y_i}= 0\\[2pt]
&\forall c\neq y_i: ~ 0\leq\alpha_{i,c}\leq C.
\end{aligned}
\begin{aligned}
\end{aligned}
\end{equation}

\section{Experiments}

\subsection{Setup}\label{sec:exprsetup}
For our experiments we use two different types of machines. Type A has 20 physical cpu cores, 128 GB of memory and a 10 GigaBit Ethernet network. Type B has 24 physical cpu cores and 386 GB of memory. On type B we ran the experiments involving CS due to the memory requirements.

Training repetitions were run on training sets with a random order of the data (note that the training set is the same in each run; only the order of points is shuffled, which can impact the DBCA algorithm). For \textit{LIBLINEAR} solvers we use the newest available version as of April 2016 with the default settings.

We implemented our solveres using OpenMP, OpenMPI, and the Python-ecosystem. In more detailed we used \cite{van2011numpy}, \cite{behnel2011cython}, and \cite{dalcin2011parallel}.

\subsection{Validation}
The following Table~\ref{tab:smallDataResultsAll} contains the complete results of the validation experiment in Section~\ref{sec:validationexpr}.

  \begin{table*}[b]
    \footnotesize
    \centering
    \caption{ {\normalsize Error on the test set and density in \% of the Shark solver (denoted S) and the proposed solver (denoted D).}\label{tab:smallDataResultsAll}}
    \begin{tabular}{ |r | r|r | r|r | r|r | r|r |}
    \hline		
    & \multicolumn{2}{c|}{\textbf{D-LLW}} & \multicolumn{2}{c|}{\textbf{S-LLW}} & \multicolumn{2}{c|}{\textbf{D-WW}} & \multicolumn{2}{c|}{\textbf{S-WW}} \\
    \cline{2-9}
    \textbf{Dataset:}  & Err. & Den. & Err. & Den. & Err. & Den. & Err. & Den. \\ 
    \hline
    \hline
    \textbf{SensIT v.(com.)} & & & & & & & & \\
    \textit{$\log(C)$: -1} & 21.34 & 100.0 & 21.34 & 100.0 & 19.88 & 100.0 & 19.88 & 100.0 \\
    \textit{    0} & 20.95 & 100.0 & 20.95 & 100.0 & 19.51 & 100.0 & 19.51 & 100.0 \\
    \textit{    1} & 20.78 & 100.0 & 20.78 & 100.0 & 19.38 & 100.0 & 19.38 & 100.0 \\

    \textbf{glass}  & & & & & & & & \\
    \textit{$\log(C)$: -1} & 66.67 & 100.0 & 66.67 & 100.0 & 38.10 & 100.0 & 38.10 & 100.0 \\
    \textit{    0} & 61.90 & 100.0 & 61.90 & 100.0 & 19.05 & 100.0 & 19.05 & 100.0 \\
    \textit{    1} & 33.33 & 100.0 & 33.33 & 100.0 & 19.05 & 100.0 & 19.05 & 100.0 \\

    \textbf{iris} & & & & & & & & \\
    \textit{$\log(C)$: -1} & 13.33 & 100.0 & 13.33 & 100.0 &  6.67 & 100.0 &  6.67 & 100.0 \\
    \textit{    0} & 26.67 & 100.0 & 26.67 & 100.0 & 13.33 & 100.0 & 13.33 & 100.0 \\
    \textit{    1} & 26.67 & 100.0 & 26.67 & 100.0 & 13.33 & 100.0 & 13.33 & 100.0 \\

    \textbf{letter} & & & & & & & & \\
    \textit{$\log(C)$: -1} & 87.04 & 100.0 & 87.04 & 100.0 & 28.25 & 100.0 & 28.26 & 100.0 \\
    \textit{    0} & 87.24 & 100.0 & 87.24 & 100.0 & 29.04 & 100.0 & 29.03 & 100.0 \\
    \textit{    1} & 61.91 & 100.0 & 87.24 & 100.0 & 28.92 & 100.0 & 28.93 & 100.0 \\

    \textbf{news20} & & & & & & & & \\
    \textit{$\log(C)$: -1} & 29.23 & 97.24 & 29.23 & 97.24 & 15.32 & 51.16 & 15.30 & 49.72 \\
    \textit{    0} & 22.97 & 97.24 & 22.97 & 97.24 & 14.80 & 44.74 & 14.80 & 42.70 \\
    \textit{    1} & 16.15 & 97.17 & 16.15 & 97.04 & 15.98 & 45.97 & 15.98 & 43.47 \\

    \textbf{rcv1} & & & & & & & & \\
    \textit{$\log(C)$: -1} & 47.96 & 78.00 & 47.96 & 78.00 & 11.31 & 26.42 & 11.31 & 23.45 \\
    \textit{    0} & 33.27 & 78.00 & 33.41 & 77.98 & 11.52 & 22.93 & 11.52 & 20.12 \\
    \textit{    1} & 12.03 & 78.00 & 12.03 & 77.98 & 12.03 & 23.05 & 12.03 & 20.06 \\

    \textbf{satimage} & & & & & & & & \\
    \textit{$\log(C)$: -1} & 26.75 & 100.0 & 26.73 & 100.0 & 15.80 & 100.0 & 15.80 & 100.0 \\
    \textit{    0} & 26.80 & 100.0 & 26.80 & 100.0 & 15.47 & 100.0 & 15.53 & 100.0 \\
    \textit{    1} & 26.90 & 100.0 & 26.90 & 100.0 & 15.96 & 100.0 & 16.00 & 100.0 \\ 

    \textbf{splice} & & & & & & & & \\
    \textit{$\log(C)$: -1} & 16.29 & 100.0 & 16.37 & 100.0 & 16.16 & 100.0 & 16.16 & 100.0 \\
    \textit{    0} & 16.09 & 100.0 & 16.15 & 100.0 & 16.37 & 100.0 & 16.28 & 100.0 \\
    \textit{    1} & 16.34 & 100.0 & 16.28 & 100.0 & 16.32 & 100.0 & 16.24 & 100.0 \\ 

    \textbf{usps}  & & & & & & & & \\
    \textit{$\log(C)$: -1} & 31.84 & 100.0 & 31.84 & 100.0 &  8.17 & 100.0 &  8.17 & 100.0 \\
    \textit{    0} & 30.09 & 100.0 & 30.04 & 100.0 &  9.37 & 100.0 &  9.37 & 100.0 \\
    \textit{    1} & 28.00 & 100.0 & 28.00 & 100.0 & 10.51 & 100.0 & 10.51 & 100.0 \\
    \hline
  \end{tabular}
  \end{table*}

  \subsection{LSHTC F1-Scores}

  The following Table~\ref{tab:lshtcF1Results} contains the F1-Scores achieved by the solvers on the LSHTC1 datasets.

  \begin{table*}[t]
    \footnotesize
    \centering
    \caption{Micro-F1 and Macro-F1 scores (in \%) as achieved by the OVR, CS, WW, and LLW solvers on the LSHTC datasets. For each solver and each metric the best result across C values is in bold font. For LLW entries with a '*' did not converge within a day of runtime.}\label{tab:lshtcF1Results}
    \begin{tabular}{ r | r r r r || r r r r }
    & \multicolumn{4}{c||}{\textbf{\textit{Micro-F1}}} & \multicolumn{4}{c}{\textbf{\textit{Macro-F1}}} \\
    \textbf{Dataset:} & \textbf{OVR} & \textbf{CS} & \textbf{WW} & \textbf{LLW} & \textbf{OVR} & \textbf{CS} & \textbf{WW} & \textbf{LLW}  \\ 
    \hline
    
    \textbf{LSHTC-small}  & & & & & & & & \\
    \textit{$\log(C)$: -3} &  7.00 & 40.26 & 27.18 & \textbf{ 7.00}   &  0.61 & 22.08 & 10.73  &  \textbf{0.61} \\
    \textit{   -2}         & 14.42 & 40.26 & 34.66 &  7.00   &  2.70 & 22.08 & 16.15  &  0.61 \\
    \textit{   -1}         & 25.46 & 40.26 & 42.41 &  7.00   &  8.72 & 22.08 & 24.71  &  0.61 \\
    \textit{    0}         & 35.47 & 44.46 & 45.43 &  7.00   & 16.42 & 26.70 & 28.75  &  0.61 \\
    \textit{    1}         & \textbf{42.41} & \textbf{45.48} & \textbf{45.59} &  7.00   & \textbf{25.09} & \textbf{28.73} & \textbf{29.15}  &  0.61 \\

    \textbf{LSHTC-large}  & & & & & & & & \\
    \textit{$\log(C)$: -3} & 11.77 & 41.35 & 33.53 &  \textbf{4.14}   &  0.88 & 25.43 & 15.05 &  \textbf{0.09} \\
    \textit{   -2}         & 14.80 & 41.52 & 39.42 &  4.14   &  1.51 & 25.41 & 20.83 &  0.09 \\
    \textit{   -1}         & 22.02 & 42.19 & 44.72 &  4.14   &  3.35 & 25.83 & 27.90 &  0.09 \\
    \textit{    0}         & 36.86 & \textbf{46.41} & \textbf{46.02} &     *   & 14.76 & 30.99 & \textbf{31.29} &     * \\    
    \textit{    1}         & \textbf{42.80} & 45.83 & 45.59 &     *   & \textbf{25.87} & \textbf{31.13} & 31.12 &     * \\

    \textbf{LSHTC-2012}  & & & & & & & & \\
    \textit{$\log(C)$: -3} & 16.34 & 50.19 & 41.98 & \textbf{ 7.37}   &  0.28 & 20.55 &  8.08 &  \textbf{0.01} \\
    \textit{   -2}         & 24.85 & 50.35 & 49.80 &  7.37   &  0.69 & 20.72 & 16.17 &  0.01 \\
    \textit{   -1}         & 39.62 & 53.86 & 55.06 &  7.37   &  2.64 & 23.76 & 25.94 &  0.01 \\
    \textit{    0}         & 52.67 & \textbf{57.33} & \textbf{55.99} &     *   & 12.46 & \textbf{32.57} & \textbf{32.06} &     * \\
    \textit{    1}         & \textbf{53.17} & 54.40 & 53.85 &     *   & \textbf{24.41} & 31.84 & 30.95 &     * \\

    \textbf{LSHTC-2011}  & & & & & & & & \\
    \textit{$\log(C)$: -3} & 12.05 & 40.91 & 31.81 & \textbf{ 3.82}   &  0.46 & 22.44 & 10.47 & \textbf{ 0.05} \\
    \textit{   -2}         & 14.15 & 40.91 & 37.86 &  3.82   &  0.62 & 22.46 & 16.48 &  0.05 \\
    \textit{   -1}         & 23.22 & 41.82 & 42.69 &  3.82   &  1.89 & 23.37 & 23.17 &  0.05 \\
    \textit{    0}         & 36.89 & \textbf{44.42} & \textbf{43.06} &     *   & 10.60 & \textbf{26.97} & \textbf{27.25} &     * \\
    \textit{    1}         & \textbf{39.99} & 42.22 & 41.86 &     *   & \textbf{21.30} & 26.31 & 26.97 &     * \\
  \end{tabular}
  \end{table*}

\section{Algorithms}
The following routine complement the main-algorithms in the paper.

\begin{minipage}[!t]{0.5\textwidth}
\vspace{-0.2cm}
\begin{algorithm}[H]
  \caption{Solving the graph 1-factorization problem. Indices start with one.
    \label{alg:matching}}
  \begin{algorithmic}[1]
	\Function{MatchClass}{$\mathcal{C}$,$c$,$r$}
	\If{$\mathcal{C}$ is even \textbf{and} $c=\mathcal{C}$}
	\State \Return $r$
	\EndIf
	\If{$c=r$}
	\If{$\mathcal{C}$ is even}
	\State \Return $\mathcal{C}$
	\Else
	\State \Return $c$
	\EndIf
	\EndIf
	\State \Return $\operatorname{mod}(2r-c,\mathcal{C}-1)$ 
\EndFunction
\end{algorithmic}
\end{algorithm}
\end{minipage}

\clearpage

\end{document}